# Convex Relaxations of Bregman Divergence Clustering

**Hao Cheng**
Department of Computing Science
University of Alberta
hcheng2@ualberta.ca

**Xinhua Zhang**
Machine Learning Research Group
National ICT Australia and ANU
xinhua.zhang@nicta.com.au

**Dale Schuurmans**
Department of Computing Science
University of Alberta
dale@cs.ualberta.ca

## Abstract

Although many convex relaxations of clustering have been proposed in the past decade, current formulations remain restricted to spherical Gaussian or discriminative models and are susceptible to imbalanced clusters. To address these shortcomings, we propose a new class of convex relaxations that can be flexibly applied to more general forms of Bregman divergence clustering. By basing these new formulations on *normalized* equivalence relations we retain additional control on relaxation quality, which allows improvement in clustering quality. We furthermore develop optimization methods that improve scalability by exploiting recent implicit matrix norm methods. In practice, we find that the new formulations are able to efficiently produce tighter clusterings that improve the accuracy of state of the art methods.

## 1 Introduction

Discovering latent class structure in data, *i.e. clustering*, is a fundamental problem in machine learning and statistics. Given data, the task is to assign each observation a latent cluster label or distribution over cluster labels. Clustering has a long history, with diverse approaches proposed. Unfortunately, computational tractability remains a fundamental challenge: standard clustering formulations are *NP*-hard (Aloise et al., 2009; Dasgupta, 2008; Arora & Kannan, 2005) and additional problem structure must be postulated before efficient solutions can be guaranteed. Fortunately, standard clustering formulations are also efficiently approximable (Kumar et al., 2004), and much work has sought practical algorithms that improve solution quality, even in lieu of theoretical bounds. In this paper we contribute a new family of convex relaxations that improve clustering quality while admitting efficient algorithms.

The techniques we propose are applicable to a variety of clustering formulations. Two of the most important paradigms for clustering are based on *generative* versus *discriminative* modeling, with generative clustering consisting of hard clustering with conditional models, hard clustering with joint models, and soft clustering with joint models. We address all but soft clustering in this paper.

Traditionally, clustering formulations have used *generative models* to discover interesting latent structure in data. Let $\mathbf{X}$ denote the observation variable and $\mathbf{Y}$ denote the latent class variable. The simplest generative approach optimizes the conditional model $P(\mathbf{X}|\mathbf{Y})$ only, with $\mathbf{Y}$ assigned to the most likely value. This is also known as *hard conditional* clustering. When $P(\mathbf{X}|\mathbf{Y})$ is Gaussian, a popular approach is hard $k$-means (MacQueen, 1967) where one alternates between optimizing $\mathbf{Y}$ and the model. Banerjee et al. (2005) extended the formulation to general exponential family forms for $P(\mathbf{X}|\mathbf{Y})$ via Bregman divergences. Although hard conditional clustering provides a standard baseline, finding global solutions in this case is intractable; efficient methods are only known when the number of clusters or the dimensionality of the space is constrained (Hansen et al., 1998; Inaba et al., 1994). Consequently, there has been significant work on developing approximations, particularly via convex relaxations that can be solved in polynomial time. For example, Zha et al. (2001) derived a convex quadratic reformulation of conditional Gaussian clustering, and Peng & Wei (2007) obtained a tighter semidefinite programming (SDP) relaxation. By analyzing the complete positivity (CP) properties of the resulting constraint, Zass & Shashua (2005) propose an approximation for Gaussian clustering based on CP factorization. These can be further extended to relaxations of normalized graph-cut clustering (Xing & Jordan, 2003; Ng et al., 2001). Unfortunately, all these relaxations are restricted to Gaussian $P(\mathbf{X}|\mathbf{Y})$, and the optimization algorithms depend heavily on the linearity of the SDP objective.

The conditional clustering approach can be extended to *hard joint* clustering by explicitly including the class prior, thus optimizing the joint likelihood $P(\mathbf{X}, \mathbf{Y})$ with the most likely $\mathbf{Y}$. Again, efficient solution methods are not generally known, leaving local approaches as the only option.

To smooth these objectives, the *soft joint* model optimizes the marginal likelihood, $P(\mathbf{X}) = \sum_Y P(\mathbf{Y})P(\mathbf{X}|\mathbf{Y})$ (Neal & Hinton, 1998; Banerjee et al., 2005), which has traditionally been tackled by expectation-maximization (EM) (Dempster et al., 1977). The EM algorithm remains susceptible to local optima however. Intensive research has been devoted to understanding properties of the Gaussian mixture model in particular (Moitra & Valiant, 2010; Kalai et al., 2010; Dasgupta & Schulman, 2007; Chaudhuri et al., 2009). Although run time can be reduced to polynomial when the number of clusters or data dimensionality is constrained, it remains exponential in these quantities jointly. A few convex relaxations for soft joint clustering models have therefore been proposed. For example, Lashkari & Golland (2007) restrict cluster centers to data points, while Nowozin & Bakir (2008) exert sparsity inducing regularization over the class priors (while still embedding an intractable subproblem). Recent spectral techniques can provably recover an approximate estimate of Gaussian mixtures in polynomial time (Hsu & Kakade, 2013; Anandkumar et al., 2012). Unfortunately, this formulation remains restricted to spherical Gaussian forms of $P(\mathbf{X}|\mathbf{Y})$.

Finally, *discriminative models* provide a distinct paradigm for clustering that can be more effective when the goal of learning is to predict labels from the observation $\mathbf{X}$, *e.g.* as in semi-supervised learning (Chapelle et al., 2006). In this approach, one maximizes the reverse conditional likelihood $P(\mathbf{Y}|\mathbf{X})$, with $\mathbf{Y}$ imputed by the most likely label. A straightforward optimization strategy can alternate between optimizing $\mathbf{Y}$ and the model, but this quickly leads to local optima. Thus, here too, convex relaxation has been a popular approximation strategy, either in the case of a large margin loss (Xu & Schuurmans, 2005) or logistic loss (Joulin & Bach, 2012; Joulin et al., 2010; Bach & Harchaoui, 2007; Guo & Schuurmans, 2007). To date, such formulations have been entirely based on SDP relaxations with *unnormalized* equivalence matrices, whose elements indicate whether two examples belong to the same cluster. Such an approach is hampered by imbalanced clustering, since the model employs no mechanism to avoid assigning all examples to a single cluster.

In this paper we present new convex relaxations for hard conditional, hard joint, and discriminative clustering. One of the key results is a tighter convex relaxation of hard generative models for Bregman divergence clustering that also accounts for cluster size. We design efficient new algorithms that optimize the resulting *nonlinear* SDPs using recent induced matrix norm techniques. By applying standard rounding methods, we observe that the resulting clustering algorithms deliver lower sum of intra-cluster divergences and more faithful alignment with class labels in practice. Finally, applying our formulation to discriminative models immediately leads to normalized equivalence relations, which automatically alleviate the problem of imbalanced cluster assignment faced by current relaxations.

## 2 Background

Following (Banerjee et al., 2005), we formulate clustering as maximum likelihood estimation in an exponential family model with a latent variable $\mathbf{Y} \in \{1, \ldots, d\}$ (the class indicator). The observed variable $\mathbf{X}$ is in $\mathbb{R}^n$, from which an *iid* sample $X = (\mathbf{x}_1, \ldots, \mathbf{x}_t)'$ has been collected.

**Generative models.** In generative modeling we parameterize the joint distribution over $(\mathbf{X}, \mathbf{Y})$ as $\mathbf{Y} \to \mathbf{X}$:

$$p(\mathbf{Y} = j) = q_j, \qquad (1)$$
$$p(\mathbf{X} = \mathbf{x}|\mathbf{Y} = j) = \exp(-D_F(\mathbf{x}, \boldsymbol{\mu}_j)) Z_j(\mathbf{x}). \qquad (2)$$

Here $\Theta := \{q_j, \boldsymbol{\mu}_j\}_{j=1}^d$ are the parameters, where $\mathbf{q} \in \Delta_d$, the $d$ dimensional simplex. We assume $P(\mathbf{X}|\mathbf{Y})$ is an exponential family model defined by the Bregman divergence $D_F$, where $F$ is a strictly convex function with gradient $f = \nabla F$ (the transfer function), such that

$$D_F(\mathbf{x}, \mathbf{y}) := F(\mathbf{x}) - F(\mathbf{y}) - \langle \mathbf{x} - \mathbf{y}, f(\mathbf{y}) \rangle. \qquad (3)$$

Here it is known that $D_F(\mathbf{x}, \mathbf{y}) = D_{F^*}(f(\mathbf{y}), f(\mathbf{x}))$, where $F^*$ is the Fenchel conjugate of $F$. Also, $f^{-1}$ is well defined by the strict convexity of $F$, and $f^{-1} = \nabla F^*$. Examples of commonly used Bregman divergences include Euclidean ($f(x) = x$), and sigmoid ($f(x) = \log \frac{x}{1-x}$).

Given data $X$, the parameters $\Theta$ can be estimated via

$$\operatorname*{argmax}_\Theta \max_Y p(X, Y|\Theta) \qquad (4)$$

or $\operatorname*{argmax}_\Theta p(X|\Theta) = \max_\Theta \sum_Y p(X, Y|\Theta), \qquad (5)$

depending on whether $Y$ is to be maximized (hard clustering) or summed out (soft clustering). Here we are letting $Y$ denote a $t \times d$ assignment matrix such that $Y_{ij} \in \{0, 1\}$ and $Y\mathbf{1} = \mathbf{1}$ (a vector of all 1's with proper dimension). If we additionally let $\Gamma = (\boldsymbol{\mu}_1, \ldots, \boldsymbol{\mu}_d)$ and $B = (\mathbf{b}_1, \ldots, \mathbf{b}_d)$, such that $\mathbf{b}_j = f(\boldsymbol{\mu}_j)$, then the conditional likelihood (2) can be rewritten over the entire data set as

$$p(X|Y) = \exp(-D_F(X, Y\Gamma)) Z(X) \qquad (6)$$
$$= \exp(-D_{F^*}(YB, f(X))) Z(X), \qquad (7)$$

where $D_F(X, Y\Gamma) := \sum_{i=1}^t D_F(X_{i:}, Y_{i:}\Gamma)$ and $D_{F^*}(YB, f(X)) := \sum_{i=1}^t D_{F^*}(Y_{i:}B, f(X_{i:}))$ are row-wise sums such that $X_{i:}$ stands for the $i$-th row of $X$.

**Discriminative models.** As an alternative, discriminative clustering uses a graphical model $\mathbf{X} \to \mathbf{Y}$, and focuses on modeling the dependence of the labels $Y$ given $X$:

$$p(Y|X; W, \mathbf{b}) = \exp(-D_{F^*}(Y, f(XW + \mathbf{1}\mathbf{b}')))Z(X),$$

where $\mathbf{b} \in \mathbb{R}^d$ is the offset for all clusters. A soft clustering model cannot be applied in this case, since $\sum_Y p(X, Y) = p(X)$. Instead, hard optimization of $Y$ leads to

$$\min_{W, \mathbf{b}, Y} D_F(XW + \mathbf{1}\mathbf{b}', f^{-1}(Y)). \qquad (8)$$

All of these problems involve a mix of discrete and continuous variables, which raises considerable challenges. Our goal is to develop convex relaxations that can be solved efficiently while leading (after rounding) to higher quality solutions than those obtained by naive local optimization.

## 3 Conditional Generative Clustering

We first consider the case of *hard conditional clustering*, where the prior **q** has been fixed to some value beforehand.

### 3.1 Case 1: Jointly Convex Bregman Divergence

First note that by using (6), the estimator (4) can be reduced to $\min_{Y,\Gamma} D_F(X, Y\Gamma)$. Here Banerjee et al. (2005) showed that for any fixed assignment $Y$ the optimal $\Gamma$ is given by $\Gamma = (Y'Y)^\dagger Y'X$, for any Bregman divergence $D_F$. Plugging the solution back into the formulation, the problem becomes $\min_Y D_F(X, Y(Y'Y)^\dagger Y'X)$. Let us introduce the *normalized equivalence matrix*

$$M = Y(Y'Y)^\dagger Y' = Y \operatorname{diag}(Y'\mathbf{1})^\dagger Y', \quad (9)$$

where $\mathcal{M}$ is the set of possibilities. It then suffices to solve

$$\min_{M \in \mathcal{M}} D_F(X, MX). \quad (10)$$

This problem remains challenging for two reasons. First, the objective is not convex in $M$, since $D_F$ is only guaranteed to be convex in its first argument. However, many Bregman divergences are *jointly* convex in both arguments; *e.g.* Mahalonobis distance, KL divergence, Bernoulli entropy, Bose-Einstein entropy, Itakura-Saito distortion, and von Neumann divergence (Wang & Schuurmans, 2003; Tsuda et al., 2004). We consider this simpler case first.

The second challenge lies in the non-convexity of the constraint set $\mathcal{M}$. Peng & Wei (2007) have shown that

$$\mathcal{M} = \{M : M = M', M^2 = M, \operatorname{tr}(M) \leq d, M_{i:} \in \Delta_t\}.$$

Since $M^2 = M$ is the source of non-convexity, its convex hull can be used to construct a convex outer approximation of $\mathcal{M}$ (note that this is *not* taking the convex hull of $\mathcal{M}$):

$$\mathcal{M}_1 := \operatorname{conv}\{M : M = M' = M^2\} \cap \{M \in \Delta_t^t : \operatorname{tr}(M) \leq d\}$$
$$= \{M : \mathbf{0} \preceq M \preceq I, \operatorname{tr}(M) \leq d, M_{i:} \in \Delta_t\},$$

where by $M \succeq \mathbf{0}$ we also encode $M = M'$. Note that $M \preceq I$ is implied by $\mathbf{0} \preceq M$ and $M_{i:} \in \Delta_t$ (*e.g.* Mirsky, 1955, Theorem 7.5.4). Conveniently, $\mathcal{M}_1$ can be relaxed further by keeping only the spectral constraints

$$\mathcal{M}_2 := \{M : \mathbf{0} \preceq M \preceq I, \operatorname{tr}(M) \leq d, M\mathbf{1} = \mathbf{1}\}.$$

Although this set $\mathcal{M}_1$ has been widely used, it is still not clear whether it is the tightest convex relaxation of $\mathcal{M}$; that is, whether $\mathcal{M}_1 = \operatorname{conv}\mathcal{M}$? With some surprise, we show that this conjecture is not true in Appendix A.

#### 3.1.1 Optimization

Assuming $D_F$ is convex in its second argument, one can easily minimize $D_F(X, MX)$ over $M \in \mathcal{M}_1$ by using the alternating direction method of multipliers (ADMM) (Boyd et al., 2010). In particular, we split the constraints into two groups: spectral and non-spectral, leading to the following augmented Lagrangian:

$$\mathcal{L}(M, Z, \Lambda) = D_F(X, MX) + \delta(M_{i:} \in \Delta_t) + \delta(Z \in \mathcal{M}_2)$$
$$- \langle \Lambda, M - Z \rangle + \frac{1}{2\mu} \|M - Z\|_F^2,$$

where $\delta(\cdot) = 0$ if $\cdot$ is true; $\infty$ otherwise. The ADMM then proceeds as follows in each iteration:

1. $M_t \leftarrow \operatorname{argmin}_M \mathcal{L}(M, Z_{t-1}, \Lambda_{t-1})$; i.e. optimize objective under non-spectral constraints.

2. $Z_t \leftarrow \operatorname{argmin}_Z \mathcal{L}(M_t, Z, \Lambda_{t-1})$; i.e. project to satisfy the spectral constraints.

3. $\Lambda_t \leftarrow \Lambda_{t-1} + \frac{1}{\mu}(Z_t - M_t)$; i.e. update the multipliers.

Note that since we constrain $M_{i:} \in \Delta_t$, the objective $D_F(X, MX)$ remains well defined in Step 1. Furthermore, since the objective decomposes row-wise, each row of $M$ can be optimized independently, which constitutes a key advantage of this scheme. Second, since Step 2 merely involves projection onto spectral constraints $\mathcal{M}_2$, a closed form solution exists based on eigen-decomposition, as established in the following lemma.

**Lemma 1.** *Let $H = I - \frac{1}{t}\mathbf{1}\mathbf{1}'$. Then*

$$\mathcal{M}_2 = \{HMH + \frac{1}{t}\mathbf{1}\mathbf{1}' : M \in \mathcal{M}_3\}, \quad (11)$$

*where* $\mathcal{M}_3 = \{M : \mathbf{0} \preceq M \preceq I, \operatorname{tr}(M) \leq d-1\}$. (12)

*Proof.* Clearly the right-hand side of (11) is contained in $\mathcal{M}_2$. Conversely, for any $M_2 \in \mathcal{M}_2$, we construct an $M \in \mathcal{M}_3$ as $M = M_2 - \frac{1}{t}\mathbf{1}\mathbf{1}'$. Note that $M_2\mathbf{1} = \mathbf{1}$ implies $\mathbf{1}/\sqrt{t}$ is an eigenvector of $M_2$ with eigenvalue 1. Therefore $M \succeq \mathbf{0}$. The rest is easy to verify. □

By Proposition 1, the problem of projecting any matrix $A$ to $\mathcal{M}_2$ can be accomplished by solving

$$\min_{Z \in \mathcal{M}_2} \|Z - A\|^2 = \min_{S \in \mathcal{M}_3} \|HSH - (A - \frac{1}{t}\mathbf{1}\mathbf{1}')\|^2.$$

Let $B = A - \frac{1}{t}\mathbf{1}\mathbf{1}'$ and $V = B - HBH$. Then $HVH = \mathbf{0}$, hence the probem reduces to solving

$$\min_{S \in \mathcal{M}_3} \|HSH - HBH - V\|^2 = \min_{S \in \mathcal{M}_3} \|HSH - HBH\|^2 + \|V\|^2.$$

Now it suffices to solve $\min_{T \in \mathcal{M}_3} \|T - HBH\|^2$ and show the optimal $T$ satisfies $HTH = T$. Suppose $HBH$ has eigenvalues $\sigma_i$ and eigenvectors $\phi_i$. Then the optimal $T$ must have eigenvalues $\mu_i$ and eigenvectors $\phi_i$ such that

$$\min_{\mu_i} \sum_i (\mu_i - \sigma_i)^2, \text{ s.t. } \mu_i \in [0, 1], \sum_i \mu_i \leq d-1. \quad (13)$$

Since $\mathbf{1}$ is an eigenvector of $HBH$ with eigenvalue 0, it is trivial that the corresponding $\mu_i$ in the optimal solution is also 0. Therefore, $T\mathbf{1} = \mathbf{0}$ and $HTH = T$. Finally the optimal $Z$ is simply given by $T + \frac{1}{t}\mathbf{1}\mathbf{1}'$.

## 3.2 Case 2: Arbitrary Bregman Divergence

When the Bregman divergence is not convex in its second argument, we require a more general treatment. The key idea we exploit is to introduce a regularizer that allows a useful form of representer theorem to be applied. In particular, we augment the negative log likelihood of $P(Y|X)$ in (7) with a regularizer on the basis $B$, weighted by the number of points in the corresponding cluster. The resulting objective can be written:

$$\min_{Y,B} D_{F^*}(YB, f(X)) + \frac{\alpha}{2} \|YB\|_F^2. \quad (14)$$

Note $B$ must be in the range of $f$. By the representer theorem, there exists a matrix $A \in \mathbb{R}^{t \times n}$ such that the optimal $B$ can be written $B = (Y'Y)^\dagger Y'A$, which yields

$$\min_{M,A} D_{F^*}(MA, f(X)) + \frac{\alpha}{2} \operatorname{tr}(A'MA), \quad (15)$$

where $M$ is defined in (9). We will work with this formulation by relaxing the domain of $M$ to $\mathcal{M}_2$. Extension to $M \in \mathcal{M}_1$ is also straightforward by ADMM.

### 3.2.1 Optimization

Although (15) does not immediately exhibit joint convexity in $M$ and $A$, a change of variable immediately leads to a convex formulation. Denote $T = MA$, then $\operatorname{Im}(T) \subseteq \operatorname{Im}(M)$ where $\operatorname{Im}(M)$ is the range of $M$. Also, denote $L(Z) := D_{F^*}(Z, f(X))$ for clarity.

**Proposition 2.** *The problem* (15) *is equivalent to*

$$\min_{M \in \mathcal{M}_3} \min_{T:\operatorname{Im}(T) \subseteq \operatorname{Im}(M)} L(T) + \frac{\alpha}{2} \operatorname{tr}(T'M^\dagger T) \quad (16)$$

$$= \min_T L(T) + \frac{\alpha}{2} \underbrace{\min_{M \in \mathcal{M}_3: \operatorname{Im}(T) \subseteq \operatorname{Im}(M)} \operatorname{tr}(T'M^\dagger T)}_{:=\Omega^2(T), \text{ with } \Omega(T) \geq 0}. \quad (17)$$

*That is, any optimal* $(M, A)$ *for* (15) *provides an optimal solution to* (16) *via* $T = MA$. *Conversely, given any optimal* $(M, T)$ *for* (16), $\operatorname{Im}(T) \subseteq \operatorname{Im}(M)$ *guarantees* $T = MA$ *for some* $A$. *Thus* $(M, A)$ *is optimal for* (15).

This proposition allows one to solve a convex problem in $T$, provided that $\Omega^2(T)$ is convex and easy to compute. Interestingly, $\Omega(T)$ has other favorable properties to exploit.

**Theorem 3.** $\Omega(T)$ *defines a norm on* $T$. $\Omega$ *and its dual norm* $\Omega_*$ *can be computed in* $O(t^3)$ *and* $O(t^2 d)$ *time resp.*[1]

With these conclusions, we can optimize (17) using a generalized conditional gradient method, accelerated by local search (Laue, 2012; Zhang et al., 2012); see Algorithm 1 (further details are given in Appendix C). At each iteration, the algorithm employs a linear approximation of $L$. The inner oracle searches for a steepest descent direction by computing a subgradient of the dual norm $\Omega_*$. Algorithm 1 is

---
[1] The same conclusion holds for $M \in \mathcal{M}_2$ (see Appendix B).

---

**Algorithm 1** Conditional gradient for optimizing (17)

1: Initialize $T_0 = \mathbf{0}$. $s_0 = 0$.
2: **for** $k = 0, 1, \ldots$ **do**
3:    Set $S_k \in \partial \Omega_*(\nabla L(T_k))$, *i.e.* find a minimizer of $\min_S \langle \nabla L(T_k), S \rangle + \frac{\alpha}{2} \Omega^2(S)$ up to scaling.
4:    Line search:
   $(a, b) := \operatorname{argmin}_{a \geq 0, b \geq 0} L(aT_k + bS_k) + \frac{\alpha}{2}(as_k + b)^2$.
5:    Set $T_{k+1} = aT_k + bS_k$, $s_{k+1} = as_k + b$.
6: **end for**

---

guaranteed to find an $\epsilon$ accurate solution to (17) in $O(1/\epsilon)$ iterations; see e.g. (Zhang et al., 2012). The optimal $M$ can then be recovered by evaluating $\Omega$ at the optimal $T$.[2]

We prove Theorem 3 in three steps.

**1. Computing $\Omega$.** Let the singular values of $T$ be $s_1 \geq \ldots \geq s_t$. Since $\Omega^2(T) = \min_{M \in \mathcal{M}_3} \operatorname{tr}(TT'M^\dagger)$, by von Neumann's trace inequality (Mirsky, 1975) the optimal $M$ must have eigenvectors equal to the left singular vectors of $T$. The minimal objective value is then $\sum_i s_i^2/\sigma_i$, where $\sigma_i$ are the eigenvalues of $M$. It suffices to solve

$$f(\mathbf{s}) := \min_{\{\sigma_i\}} \sum_{i=1}^{t} \frac{s_i^2}{\sigma_i}, \text{ s.t. } \sigma_i \in [0, 1], \sum_{i=1}^{t} \sigma_i \leq d-1 \quad (18)$$

$$= \min_{\sigma_i \in [0,1]} \max_{\lambda \geq 0} \sum_{i=1}^{t} \frac{s_i^2}{\sigma_i} + \lambda \left(1 - d + \sum_{i=1}^{t} \sigma_i\right) \quad (19)$$

$$= \max_{\lambda \geq 0} \left\{ \lambda(1-d) + \min_{\sigma_i \in [0,1]} \sum_{i=1}^{t} \left(\frac{s_i^2}{\sigma_i} + \lambda \sigma_i\right) \right\}. \quad (20)$$

Fixing $\lambda$, the optimal $\sigma_i$ is attained at $\sigma_i(\lambda) = \frac{s_i}{\sqrt{\lambda}}$ if $\lambda \geq s_i^2$, and 1 if $\lambda < s_i^2$. Note that $\sigma_i(\lambda)$ decreases monotonically for $\lambda \geq s_t^2$, hence we only need to find a $\lambda$ that satisfies $\sum_{i=1}^{t} \sigma_i(\lambda) = d - 1$, since the constraint $\sum_i \sigma_i \leq d-1$ must be equality at the optimum. This only requires a line search over $\lambda$, which can be conducted efficiently as follows. Suppose the optimal $\lambda$ lies in $[s_k^2, s_{k+1}^2]$. Then $\sigma_i(\lambda) = 1$ for all $i \leq k$ and $\sigma_i(\lambda) = s_i/\sqrt{\lambda}$ for all $i > k$. So $k + \frac{1}{\sqrt{\lambda}} \sum_{i=k+1}^{t} s_i = d - 1$, hence

$$\sqrt{\lambda} = \frac{1}{d-1-k} \sum_{i=k+1}^{t} s_i \in [s_k, s_{k+1}] \Rightarrow \begin{cases} k + \frac{\sum_{i=k+1}^{t} s_i}{s_k} \leq d-1 \\ k + \frac{\sum_{i=k+1}^{t} s_i}{s_{k+1}} \geq d-1. \end{cases}$$

Now note there must be a $k$ satisfying these two conditions. Since both $k + \frac{1}{s_k} \sum_{i=k+1}^{t} s_i$ and $k + \frac{1}{s_{k+1}} \sum_{i=k+1}^{t} s_i$ grow monotonically in $k$, the smallest $k$ that satisfies the second condition must also satisfy the first condition. Hence the optimal solution is $\sigma_i = 1$ for all $i \leq k$, and $\sigma_i = (d-1-k)s_i / \sum_{i=k+1}^{t} s_i$ for $i > k$.

---
[2] This solution is valid since (16) minimizes over $M$ and $T$. If the problem were $\min_T \max_M$ instead, the optimal $M$ could not be generally recovered by maximizing $M$ for fixed optimal $T$.

**Algorithm 2** Compute $f(\mathbf{s})$ with given $d$.

1: **for** $k = 0, 1, \ldots, d-2$ **do**
2:    **if** $\sum_{i=k+1}^{t} s_i \geq (d-1-k)s_{k+1}$ **then** break
3: **end for**
4: **Return** $f(\mathbf{s}) = \sum_{i=1}^{k} s_i^2 + \frac{1}{d-1-k}\left(\sum_{i=k+1}^{t} s_i\right)^2$.

---

The algorithm for evaluating $f(\mathbf{s}) = \Omega^2(T)$ is given in Algorithm 2. The 'if' condition in step 2 must be met when $k = d-2$. The computational cost is dominated by a full SVD of $T$, and fortunately our method needs to compute $\Omega(T)$ only once at the optimal $T$.

**2. $\Omega$ is a norm.** Note that $\Omega(T)$ depends only on the singular values of $T$. So it suffices to show that $\kappa(\mathbf{s}) := \sqrt{f(\mathbf{s})}$ is a symmetric gauge (Horn & Johnson, 1985, Theorem 3.5.18), where $f(\mathbf{s})$ is defined in (18). Clearly $\kappa(\mathbf{s})$ is permutation invariant, $\kappa(a\mathbf{s}) = |a|\kappa(\mathbf{s})$ for all $a \in \mathbb{R}$, and $\kappa(\mathbf{s}) = 0$ iff $\mathbf{s} = \mathbf{0}$. So it suffices to prove the triangle inequality for $\kappa(\mathbf{s})$. For any $\mathbf{s}_1$ and $\mathbf{s}_2$, let $t_1 = \kappa(\mathbf{s}_1)$ and $t_2 = \kappa(\mathbf{s}_2)$. Then $\kappa(\frac{\mathbf{s}_1}{t_1}) = \kappa(\frac{\mathbf{s}_2}{t_2}) = 1$, and

$$\frac{\mathbf{s}_1 + \mathbf{s}_2}{t_1 + t_2} = \frac{t_1}{t_1 + t_2}\frac{\mathbf{s}_1}{t_1} + \frac{t_2}{t_1 + t_2}\frac{\mathbf{s}_2}{t_2}. \quad (21)$$

Note $f(\mathbf{s})$ is convex because $\sum_i s_i^2/\sigma_i$ is jointly convex in $(\mathbf{s}, \boldsymbol{\sigma})$, and $f(\mathbf{s})$ just minimizes out $\boldsymbol{\sigma}$. So the sub-level set at level 1 for $f$ (and $\kappa$) is convex. Therefore by (21), $\kappa((\mathbf{s}_1 + \mathbf{s}_2)/(t_1 + t_2)) \leq 1$, and so $\kappa(\mathbf{s}_1 + \mathbf{s}_2) \leq t_1 + t_2 = \kappa(\mathbf{s}_1) + \kappa(\mathbf{s}_2)$. The claim follows.

**3. Compute the subgradient of $\Omega_*$.** Given a matrix $R$, the dual norm is $\Omega_*(R) = \max_{T:\Omega(T)\leq 1} \text{tr}(R'T)$. Let the SVD of $R$ be $R = U \text{diag}\{r_1, \ldots, r_t\}V'$, where $r_1 \geq \ldots \geq r_t$. Since $\Omega$ is defined via the singular values of $T$, again by von Neumann's trace inequality the maximum is attained when the left and right singular values of $T$ are $U$ and $V$, respectively. Then $\Omega_*(R) = \max_{\mathbf{s}:f(\mathbf{s})\leq 1} \mathbf{r}'\mathbf{s}$, which by (18) is equivalent to

$$\max_{\mathbf{s}, \boldsymbol{\sigma}} \mathbf{r}'\mathbf{s}, \text{ s.t. } \sigma_i \in [0,1], \sum_{i=1}^{t}\sigma_i \leq d-1, \sum_{i=1}^{t}\frac{s_i^2}{\sigma_i} \leq 1. \quad (22)$$

Using the Cauchy-Schwarz inequality, we have

$$\mathbf{r}'\mathbf{s} = \sum_{i=1}^{t}\frac{s_i}{\sqrt{\sigma_i}} \cdot r_i\sqrt{\sigma_i} \leq \left(\sum_{i=1}^{t}\frac{s_i^2}{\sigma_i}\right)^{1/2}\left(\sum_{i=1}^{t}r_i^2\sigma_i\right)^{1/2}$$

$$\leq \left(\sum_{i=1}^{t}r_i^2\sigma_i\right)^{1/2} \leq \|(r_1, r_2, \ldots, r_{d-1})'\|. \quad (23)$$

where the last two inequalities use the constraints in (22). The equalities can all be attained by setting $s_i = r_i/\|(r_1, r_2, \ldots, r_{d-1})'\|$ and $\sigma_i = 1$ for $i \leq d-1$, and $s_i = 0$ and $\sigma_i = 0$ for $i \geq d$. Clearly $U \text{diag}(\mathbf{s})V'$ is a subgradient of $\Omega_*$ at $R$. Evaluating the dual norm is inexpensive, since it requires only the top $d-1$ singular values of $R$.

## 4 Discriminative Clustering

Although generative models can often reveal useful latent structure in data, many problems such as semi-supervised learning and multiple instance learning are more concerned with accurate label prediction. In such settings, discriminative models $\mathbf{X} \to \mathbf{Y}$ can often be more effective (Joulin & Bach, 2012; Bach & Harchaoui, 2007; Guo & Schuurmans, 2007; Xu & Schuurmans, 2005).

Before attempting a convex relaxation for the discriminative model (8), it is important to recognize that a plain optimization over $(W, \mathbf{b}, Y)$ will lead to vacuous solutions, where all examples are assigned to a single cluster $j$ and $b_j = \infty$. A common solution is to add a regularizer on $Y$ to enforce a more balanced cluster distribution. Note that this situation is opposite of generative clustering, where one must upper bound $d$, since otherwise the joint likelihood would be trivially maximized by assigning each data point to its own cluster.

For discriminative clustering, we consider a special case $F(\mathbf{x}) = \log \sum_i \exp(x_i)$, i.e. where the transfer $\nabla F$ is sigmoidal (Joulin & Bach, 2012). A natural choice of regularizer on $Y$ is the entropy of cluster sizes, i.e. $-h(Y'\mathbf{1})$ where $h(\mathbf{x}) = \sum_i x_i \log x_i$. In this setting, we derive a convex relaxation for discriminative clustering that uses the normalized equivalence matrix.

By adding value regularization $\|WY'\|^2$ to (8), one obtains

$$\min_{W,\mathbf{b},Y} \frac{1}{t}D_F(XW + \mathbf{1b}', f^{-1}(Y)) + \frac{\gamma}{2}\|WY'\|^2 + h(Y'\mathbf{1})$$

$$= \min_{W,\mathbf{b},Y} \frac{1}{t}F(XW + \mathbf{1b}') - \frac{1}{t}\text{tr}((XW + \mathbf{1b}')Y')$$

$$- \frac{1}{t}F(Y) + \frac{\gamma}{2}\|WY'\|^2 + h(Y'\mathbf{1})$$

$$= \min_{W,\mathbf{b},Y} \max_{\Lambda:\Lambda_i:\in\Delta} -\frac{1}{t}F^*(\Lambda) + \frac{1}{t}\text{tr}(\Lambda'(XW + \mathbf{1b}'))$$

$$- \frac{1}{t}F(Y) - \text{tr}((XW + \mathbf{1b}')Y') + \frac{\gamma}{2}\|WY'\|^2 + h(Y'\mathbf{1})$$

$$= \min_{W,\mathbf{b},Y} \max_{\Omega:\Omega_i:\in\Delta} -\frac{1}{t}F^*(\Omega Y) + \frac{1}{t}\text{tr}(Y'\Omega'(XW + \mathbf{1b}'))$$

$$- \frac{1}{t}F(Y) - \frac{1}{t}\text{tr}((XW + \mathbf{1b}')Y') + \frac{\gamma}{2}\|WY'\|^2 + h(Y'\mathbf{1}).$$

Here, the second step follows from Fenchel's identity $F(\mathbf{x}) = \max_{\mathbf{z}\in\text{dom }F^*} \mathbf{x}'\mathbf{z} - F^*(\mathbf{z})$, where dom denotes the effective domain of a convex function. The last step involves a change of variable, $\Lambda = \Omega Y$, and converted the constraints on $\Lambda$ to $\Omega_i: \in \Delta$ (Guo & Schuurmans, 2007). By taking the gradient with respect to $W$ and $\mathbf{b}$, one obtains

$$W = \frac{1}{t}X'(I - \Omega)Y(Y'Y)^\dagger, \text{ and } \Omega'\mathbf{1} = \mathbf{1}. \quad (24)$$

Note that $-\frac{1}{t}F^*(\Omega Y) + h(Y'\mathbf{1}) \leq -\frac{1}{t}F^*(\Omega) + c_0$ where $c_0$ is some constant (Joulin & Bach, 2012, Eq 3). Using

(24) and the fact that $F(Y)$ is a constant, one can upper bound the objective by

$$\min_{M\in\mathcal{M}} \max_{\Omega:\Omega_{i:}\in\Delta,\Omega'\mathbf{1}=\mathbf{1}} -\frac{1}{t}F^*(\Omega) - \frac{1}{2\gamma t^2}\|X'(I-\Omega)M\|^2. \quad (25)$$

Importantly, this formulation is expressed completely in terms of the normalized equivalence matrix $M$, which constitutes a significant advantage over (Joulin & Bach, 2012; Guo & Schuurmans, 2007). Rather than resort to the proximal gradient method to solve for $\Omega$ given $M$ (Joulin & Bach, 2012), which is slow in practice, we can harness the power of second order solvers like L-BFGS by dualizing the problem back to the primal form, which leads to an unconstrained problem. This reformulation also sheds light on the nature of the relaxation (25).

Fixing $M \in \mathcal{M}$, we add a Lagrange multiplier $\boldsymbol{\tau} \in \mathbb{R}^t$ to enforce $\Omega'\mathbf{1} = \mathbf{1}$. By introducing the change of variable $\Psi = I - \Omega$, the optimization over $\Omega$ becomes equivalent to

$$\min_{\Psi\leq I:\Psi\mathbf{1}=\mathbf{0}} \frac{1}{t}F^*(I-\Psi) + \frac{1}{2\gamma t^2}\|X'\Psi M\|^2 + \frac{1}{t}\boldsymbol{\tau}'\Psi\mathbf{1}. \quad (26)$$

The tool we use for dualization is provided by the following lemma.

**Lemma 4. (Borwein & Lewis, 2000, Theorem 3.3.5)** *Let $J$ and $G$ be convex functions, and $A$ a linear transform. Suppose $A \operatorname{dom} J$ has nonempty intersection with $\{\mathbf{x} \in \operatorname{dom} G^* : G^*$ is continuous at $\mathbf{x}\}$. Then*

$$\min_{\mathbf{x}} J(\mathbf{x}) + G(A\mathbf{x}) = \max_{\mathbf{y}} -J^*(-A'\mathbf{y}) - G^*(\mathbf{y}). \quad (27)$$

To apply Lemma 4 to (26), choose the linear transform $A$ to be $\Psi \mapsto \frac{1}{t}X'\Psi M$, $G(\Psi) = \frac{1}{2\gamma}\operatorname{tr}(\Psi M^\dagger \Psi')$,[3] and $J(\Psi) = \frac{1}{t}F^*(I-\Psi) + \frac{1}{t}\boldsymbol{\tau}'\Psi\mathbf{1}$ over $\Psi\mathbf{1} = \mathbf{0}$ and $\Psi \leq I$ (elementwise). Then the problem (26) becomes equivalent to

$$\min_{M,\boldsymbol{\tau},\Upsilon\in\mathbb{R}^{t\times n}} \frac{1}{t}\sum_i [F(\tfrac{1}{t}X_{i:}\Upsilon'M + \boldsymbol{\tau}') - (\tfrac{1}{t}X_{i:}\Upsilon'M_{:i} + \tau_i)]$$
$$+ \frac{\gamma}{2}\operatorname{tr}(\Upsilon'M\Upsilon). \quad (28)$$

Note that $F = g$ can be interpreted as a soft max, hence the result is related to the typical max-margin style model. The loss of each example $i$ is the soft max of $X_{i:}\Upsilon'M + \boldsymbol{\tau}'$ (a row vector) minus $X_{i:}\Upsilon'M_{:i} + \tau_i$. Here $\tau_i$ is an offset associated with each training example (cf. $b_j$ for each cluster).

### 4.1 Optimization

The most straightforward method for optimizing (28) is to treat it as a convex function of $M$, whose gradient and objective value can be evaluated by minimizing out $\Upsilon$ and $\boldsymbol{\tau}$.

---

[3] Since $M^2 = M$ for $M \in \mathcal{M}$, (26) can also be recovered by setting $G(\Psi) = \frac{1}{2\gamma}\operatorname{tr}(\Psi\Psi')$. However, to reformulate the problem into (29), which is the key to efficient optimization, it is crucial to include $M^\dagger$ in $G$.

Since both $\Upsilon$ and $\boldsymbol{\tau}$ are unconstrained, this can be easily accomplished by quasi-Newton methods like L-BFGS. Interestingly, thanks to the structure of the problem, we can optimize (28) even more efficiently by applying the same change of variable as in §3.2.1. Letting $V = M\Upsilon \in \mathbb{R}^{t\times n}$ and constraining $M$ to $\mathcal{M}_3$, the problem (28) becomes

$$\min_{V,\boldsymbol{\tau}} \frac{\gamma}{2}\Omega^2(V) + \frac{1}{t}\sum_i [F(\tfrac{1}{t}X_{i:}V' + \boldsymbol{\tau}') - (\tfrac{1}{t}X_{i:}V'_{i:} + \tau_i)]. \quad (29)$$

This objective again absorbs the spectral constraints on $M$ into the norm $\Omega$, and can be readily solved by generalized conditional gradient in Algorithm 1. The extension to $M \in \mathcal{M}_2$ is also immediate.

## 5 Joint Generative Clustering

In all models considered so far, we have ignored the cluster prior $\mathbf{q}$. This quantity is often useful in practice for inference at the cluster level, and can often be effectively learned by joint generative models. In this section, we extend our convex relaxation technique to this setting.

Assume a multinomial distribution over cluster prior parameterized by $\mathbf{w} \in \mathbb{R}^d$: $p(\mathbf{Y} = j) = \exp(w_j - g(\mathbf{w}))$ where $g(\mathbf{w}) = \log\sum_i \exp(x_i)$. Then by (1) and (7), the negative log joint likelihood is: $-\mathbf{1}'Y\mathbf{w} + tg(\mathbf{w}) + L(YB) + \text{const}$. As above, one can add regularizers on $\mathbf{w}$ and $B$, as well as an entropic regularizer $h(Y'\mathbf{1})$ to encourage cluster diversity, yielding:

$$\min_{\mathbf{w},B,Y} -\frac{1}{t}\mathbf{1}'Y\mathbf{w} + g(\mathbf{w}) + \frac{\beta}{2}\|Y\mathbf{w}\|^2 + h(Y'\mathbf{1}) \quad (30)$$
$$+ \frac{1}{t}L(YB) + \frac{\alpha}{2}\|YB\|_F^2.$$

This formulation can be convexified in terms of $M$ by using the same techniques as §4 and §3.2, respectively. In particular, consider the prior $p(Y)$ as a discriminative model $Z \to Y$, where $Z$ can only take a constant scalar value 1. Then treating $Z$ as the $X$ in §4, it is easy to show that the first line of (30) can be relaxed into (ignoring the offset $\boldsymbol{\tau}$):

$$\min_{\mathbf{s}\in\mathbb{R}^t} \frac{\beta}{2}\operatorname{tr}(\mathbf{s}'M\mathbf{s}) - \frac{1}{t}\mathbf{1}'M\mathbf{s} + g\left(\frac{1}{t}M\mathbf{s}\right). \quad (31)$$

Finally by applying the same technique that converted (14) to (15) in conditional model, one can reformulate (30) into:

$$\min_{A,M,\mathbf{s}} \frac{\beta}{2}\operatorname{tr}(\mathbf{s}'M\mathbf{s}) - \frac{1}{t}\mathbf{1}'M\mathbf{s} + g(\tfrac{1}{t}M\mathbf{s}) \quad (32)$$
$$+ \frac{1}{t}L(MA) + \frac{\alpha}{2}\operatorname{tr}(A'MA).$$

To optimize this formulation, let $\mathbf{u} = M\mathbf{s} \in \mathbb{R}^t$ and $T =$

| Data set | $t$ | $n$ | $d$ | Data set | $t$ | $n$ | $d$ |
|---|---|---|---|---|---|---|---|
| Yale | 165 | 1024 | 15 | Diabetes | 768 | 8 | 2 |
| ORL | 400 | 1024 | 40 | Heart | 270 | 13 | 2 |
| E-mail | 1000 | 57 | 2 | Breast | 699 | 9 | 2 |
| Balance | 625 | 4 | 2 | | | | |

Table 1: Properties of the data sets used in the experiments.

$MA \in \mathbb{R}^{t \times n}$. Then with $M \in \mathcal{M}_3$, (32) becomes

$$\min_{\mathbf{u},T} g\left(\frac{\mathbf{u}}{t}\right) - \frac{1}{t}\mathbf{1}'\mathbf{u} + \frac{1}{t}L(T) + \min_{M \in \mathcal{M}_3} \frac{\beta}{2}\mathbf{u}'M^{\dagger}\mathbf{u} + \frac{\alpha}{2}\text{tr}(T'M^{\dagger}T)$$

$$= \min_{\mathbf{u},T} g\left(\frac{\mathbf{u}}{t}\right) - \frac{1}{t}\mathbf{1}'\mathbf{u} + \frac{1}{t}L(T) + \frac{1}{2}\Omega^2([\sqrt{\beta}\mathbf{u}, \sqrt{\alpha}T]), \quad (33)$$

which can be solved by the methods outlined above.

## 6 Experimental Evaluation

We evaluated the proposed convex relaxations for the three models developed in this paper: conditional (jointly convex or arbitrary Bregman divergence), joint, and discriminative.

**Data sets.** We used seven labeled data sets for these experiments. Five of them are from the UCI repository (Frank & Asuncion, 2010): Balance, Breast Cancer, Diabetes, Heart, and Spam E-mail. The two others are multiclass face data sets: ORL[4] and Yale[5]. We down-sampled Spam-Email to 1000 points while preserving the class ratio. The properties of these data sets are summarized in Table 1, giving the values of $t$, $n$, and $d$. We shifted all features to be nonnegative so that all transfer functions can be applied. Finally the features were normalized to unit variance.

**Transfer functions.** For all generative models, we tested two transfer functions: linear and sigmoid.

**Parameters settings.** To closely approximate the original objective without creating numerical difficulty, we chose all the regularization parameters $\alpha$, $\beta$ and $\gamma$ to be reasonably small $\alpha \in \{10^{-5}, 10^{-9}\}$, $\beta \in \{10^{-5}, 10^{-9}\}$, $\gamma \in \{10^{-6}, 10^{-9}\}$ and report the experimental results for the choices that obtain highest accuracy. However, the results were not sensitive to these values.

### 6.1 Conditional: Jointly Convex Bregman Divergence

**Algorithms.** Our method (cvxCondJC) first minimizes $D_F(X, MX)$ as in (10), but over $M \in \mathcal{M}_1$. The optimal $M$ is then rounded to a hard cluster assignment via spectral clustering (SC rounding, Shi & Malik, 2000). The result is further used to initialize a local re-optimization using the *original* objective $D_F(X, Y\Gamma)$. Since $k$-class spectral clustering involves a $k$-means algorithm, with random elements, this was repeated 10 times and variance reported.

[4] cl.cam.ac.uk/research/dtg/attarchive/facedatabase.html
[5] http://cvc.yale.edu/projects/yalefaces/yalefaces.html

|  | cvxCondJC +SC rounding | cvxCondJC +SC+re-opt | altCondJC |
|---|---|---|---|
| | | Spam E-mail | |
| lin_obj($\times 10^2$) | 9.4± 0.1 | **9.3**± 0.0 | 9.3± 0.0 |
| lin_acc(%) | 71.5±11.6 | **76.3**±13.6 | 75.1±12.6 |
| sigm_obj($\times 10^3$) | 7.8± 0.1 | 7.7± 0.1 | 7.7± 0.1 |
| sigm_acc(%) | 75.1±12.0 | **80.0**± 9.4 | 76.0± 7.2 |
| | | ORL | |
| lin_obj($\times 10^3$) | 3.3± 0.1 | **2.0**± 0.0 | 2.1± 0.0 |
| lin_acc(%) | **57.0**± 3.5 | 55.4± 2.9 | 40.6± 2.3 |
| sigm_obj($\times 10^2$) | 3.8± 0.1 | **3.5**± 0.1 | 3.7± 0.1 |
| sigm_acc(%) | 57.8± 3.6 | **58.2**± 4.1 | 48.2± 3.0 |
| | | Yale | |
| lin_obj($\times 10^1$) | 5.6± 0.1 | **5.5**± 0.0 | 5.8± 0.1 |
| lin_acc(%) | 46.8± 1.7 | **47.0**± 2.1 | 44.5± 4.2 |
| sigm_obj($\times 10^2$) | 9.6± 0.4 | **9.2**± 0.1 | 9.6± 0.3 |
| sigm_acc(%) | 49.9± 2.1 | **51.5**± 2.1 | 46.6± 4.1 |
| | | Balance | |
| lin_obj($\times 10^1$) | 7.2± 0.0 | **7.1**± 0.0 | 7.2± 0.0 |
| lin_acc(%) | 57.1± 6.9 | **57.3**± 7.1 | 54.2± 4.6 |
| sigm_obj($\times 10^2$) | 5.0± 0.3 | **3.9**± 0.0 | 4.0± 0.0 |
| sigm_acc(%) | 49.3± 5.1 | **50.5**± 5.1 | 49.4± 4.3 |
| | | Breast Cancer | |
| lin_obj($\times 10^2$) | 1.8± 0.2 | **1.6**± 0.0 | 1.7± 0.0 |
| lin_acc(%) | 72.5±12.7 | **84.7**± 8.8 | 78.7±10.4 |
| sigm_obj($\times 10^2$) | **8.5**± 0.2 | 8.5± 0.1 | 8.5± 0.1 |
| sigm_acc(%) | 72.4±13.7 | **72.5**±13.7 | 70.6±11.6 |
| | | Diabetes | |
| lin_obj($\times 10^2$) | **2.0**± 0.1 | 2.0± 0.0 | 2.0± 0.0 |
| lin_acc(%) | 57.1± 0.5 | 58.5± 0.0 | 58.5± 0.1 |
| sigm_obj($\times 10^3$) | 1.2± 0.1 | **1.1**± 0.0 | 1.1± 0.0 |
| sigm_acc(%) | **58.8**± 3.9 | 58.2± 0.1 | 58.0± 0.6 |
| | | Heart | |
| lin_obj($\times 10^2$) | 1.3± 0.0 | 1.3± 0.0 | 1.3± 0.0 |
| lin_acc(%) | **68.1**±10.0 | 65.6± 7.8 | 65.4± 5.0 |
| sigm_obj($\times 10^2$) | 7.5± 0.2 | **7.2**± 0.2 | 7.2± 0.2 |
| sigm_acc(%) | 63.4± 5.9 | **64.9**± 6.6 | 64.4± 7.8 |

Table 2: Experimental results for the conditional model with jointly convex Bregman divergences. Here "lin" and "sigm" refer to linear and sigmoid transfers respectively. Best results in **bold**.

We compared our algorithm with altCondJC (hard EM), which optimizes $D_F(X, Y\Gamma)$ by alternating, with $Y$ reinitialized randomly 30 times.

**Results.** In Table 2, the first and third rows of each block gives the optimal value of $D_F(X, Y\Gamma)$ found by altCondJC, and by cvxCondJC (both after SC rounding and re-optimization). The second and fourth lines give the highest accuracy among all possible matchings between the clusters and ground truth labels. Across all data sets and transfer functions, cvxCondJC with SC rounding and re-optimization finds a lower objective value and higher accuracy than altCondJC. In addition, although the objective achieved after rounding might be higher than that of altCondJC, the accuracy is usually comparable. Overall, the final clustering found by cvxCondJC is superior to randomized local optimization.

| | cvxCond +SC rounding | cvxCond +SC rounding & re-opt | altCond |
|---|---|---|---|
| | | Spam E-mail | |
| lin_obj($\times 10^2$) | **9.3**± 0.1 | **9.3**± 0.0 | **9.3**± 0.0 |
| lin_acc(%) | 75.0± 9.0 | **79.8**±10.2 | 73.9±13.3 |
| sigm_obj($\times 10^3$) | 8.0± 0.2 | **7.7**± 0.1 | **7.7**± 0.1 |
| sigm_acc(%) | 64.8±12.5 | **78.7**± 7.8 | 75.3± 5.5 |
| | | ORL | |
| lin_obj($\times 10^3$) | 2.7± 0.1 | **2.0**± 0.0 | 2.1± 0.0 |
| lin_acc(%) | **62.6**± 3.0 | 59.4± 2.4 | 40.1± 2.3 |
| sigm_obj($\times 10^2$) | 4.0± 0.1 | **3.4**± 0.0 | 3.7± 0.1 |
| sigm_acc(%) | **60.1**± 6.1 | 60.0± 4.9 | 48.6± 2.7 |
| | | Yale | |
| lin_obj($\times 10^1$) | 6.1± 0.2 | **5.7**± 0.1 | 5.8± 0.1 |
| lin_acc(%) | 43.3± 3.2 | **45.2**± 3.2 | 44.4± 4.0 |
| sigm_obj($\times 10^2$) | 10.3± 0.2 | **9.3**± 0.1 | 9.5± 0.2 |
| sigm_acc(%) | 46.6± 2.6 | **51.1**± 2.7 | 46.2± 3.0 |
| | | Balance | |
| lin_obj($\times 10^1$) | 8.0± 0.4 | **7.1**± 0.0 | **7.1**± 0.0 |
| lin_acc(%) | 57.1± 6.9 | **57.3**± 7.1 | 55.5± 5.1 |
| sigm_obj($\times 10^2$) | 4.0± 0.0 | **3.9**± 0.0 | 4.0± 0.1 |
| sigm_acc(%) | **54.1**± 8.3 | 53.0± 6.0 | 50.9± 5.2 |
| | | Breast Cancer | |
| lin_obj($\times 10^2$) | 1.7± 0.1 | **1.6**± 0.0 | 1.7± 0.0 |
| lin_acc(%) | 75.4±13.3 | **85.8**± 6.6 | 78.7±10.9 |
| sigm_obj($\times 10^2$) | 8.8± 0.2 | **8.5**± 0.1 | 8.6± 0.2 |
| sigm_acc(%) | 66.8± 8.4 | **72.3**±12.5 | 70.3±11.0 |
| | | Diabetes | |
| lin_obj($\times 10^2$) | **2.0**± 0.0 | **2.0**± 0.0 | **2.0**± 0.0 |
| lin_acc(%) | 58.1± 0.6 | **58.3**± 0.0 | 58.2± 0.1 |
| sigm_obj($\times 10^3$) | 1.2± 0.1 | 1.1± 0.0 | **1.0**± 0.0 |
| sigm_acc(%) | 54.7± 3.0 | **58.2**± 0.2 | 58.1± 0.5 |
| | | Heart | |
| lin_obj($\times 10^2$) | **1.3**± 0.0 | **1.3**± 0.0 | **1.3**± 0.0 |
| lin_acc(%) | **69.4**± 9.3 | 67.0± 5.5 | 66.1± 5.2 |
| sigm_obj($\times 10^2$) | 7.2± 0.1 | **7.1**± 0.1 | 7.3± 0.2 |
| sigm_acc(%) | **66.9**±10.7 | 64.9± 8.2 | 65.8± 6.3 |

Table 3: Experimental results for the conditional model with arbitrary Bregman divergences. Best results shown in **bold**.

| | cvxDisc | JB | GS |
|---|---|---|---|
| | | Spam E-mail | |
| run time ($\times 10^4$s) | **0.005** | 0.651 | 2.148 |
| obj w/ SC rounding ($\times 10^3$) | **8.0**± 0.2 | 8.7± 0.0 | 8.2± 0.2 |
| obj w/ SC + re-opt ($\times 10^3$) | **7.6**± 0.0 | 7.9± 0.2 | **7.6**± 0.0 |
| acc w/ SC rounding (%) | **69.9**±14.3 | 60.7± 0.1 | 62.8± 9.2 |
| acc w/ SC + re-opt (%) | **83.5**± 7.8 | 61.3± 9.2 | 81.4± 5.6 |
| | | ORL | |
| run time ($\times 10^4$s) | **0.080** | 0.695 | 6.372 |
| obj w/ SC rounding ($\times 10^2$) | 4.1± 0.1 | 7.1± 0.0 | **3.6**± 0.0 |
| obj w/ SC + re-opt ($\times 10^3$) | **3.5**± 0.0 | 3.8± 0.1 | 3.6± 0.0 |
| acc w/ SC rounding (%) | **59.4**± 2.7 | 20.0± 1.1 | 54.6± 2.1 |
| acc w/ SC + re-opt (%) | **59.5**± 2.8 | 45.2± 2.5 | 54.6± 2.4 |
| | | Yale | |
| run time ($\times 10^3$s) | **0.050** | 0.648 | 6.745 |
| obj w/ SC rounding ($\times 10^3$) | **8.6**± 0.2 | 13.2± 0.0 | 10.2± 0.3 |
| obj w/ SC + re-opt ($\times 10^3$) | **7.6**± 0.1 | 8.3± 0.1 | 7.8± 0.3 |
| acc w/ SC rounding (%) | **44.3**± 2.5 | 16.2± 0.6 | 33.8± 3.6 |
| acc w/ SC + re-opt (%) | **46.1**± 2.9 | 34.1± 2.6 | 42.4± 2.7 |
| | | Balance | |
| run time ($\times 10^4$s) | **0.004** | 0.155 | 0.078 |
| obj w/ SC rounding ($\times 10^2$) | 5.1± 0.0 | 6.1± 0.0 | **4.9**± 0.1 |
| obj w/ SC + re-opt ($\times 10^2$) | **3.9**± 0.0 | 4.5± 0.0 | 4.1± 0.2 |
| acc w/ SC rounding (%) | **62.0**± 2.3 | 47.0± 1.8 | 46.5± 6.3 |
| acc w/ SC + re-opt (%) | 58.7± 0.0 | **62.3**± 1.8 | 52.2± 5.2 |
| | | Breast Cancer | |
| run time ($\times 10^4$s) | **0.006** | 0.479 | 1.758 |
| obj w/ SC rounding ($\times 10^2$) | **8.5**± 0.0 | 10.0± 0.0 | 9.1± 0.2 |
| obj w/ SC + re-opt ($\times 10^2$) | **8.4**± 0.0 | 8.7± 0.3 | **8.4**± 0.1 |
| acc w/ SC rounding (%) | **79.8**±15.7 | 60.4± 3.6 | 72.3±10.3 |
| acc w/ SC + re-opt (%) | 80.7±12.5 | 60.0± 4.2 | **84.4**± 8.8 |
| | | Diabetes | |
| run time ($\times 10^4$s) | **0.012** | 1.722 | 2.731 |
| obj w/ SC rounding ($\times 10^3$) | **1.2**± 0.1 | 1.4± 0.0 | 1.3± 0.1 |
| obj w/ SC + re-opt ($\times 10^3$) | **1.1**± 0.0 | **1.1**± 0.0 | **1.1**± 0.0 |
| acc w/ SC rounding (%) | 53.5± 3.1 | **64.8**± 0.0 | 56.6± 4.2 |
| acc w/ SC + re-opt (%) | 58.3± 0.2 | **58.6**± 0.0 | 58.3± 0.2 |
| | | Heart | |
| run time ($\times 10^4$s) | **0.001** | 0.212 | 6.848 |
| obj w/ SC rounding ($\times 10^2$) | **7.6**± 0.4 | 8.6± 0.0 | 7.7± 0.4 |
| obj w/ SC + re-opt ($\times 10^3$) | **7.3**± 0.3 | 7.9± 0.0 | **7.3**± 0.2 |
| acc w/ SC rounding (%) | 61.7± 5.8 | 55.2± 0.0 | **64.4**± 9.5 |
| acc w/ SC + re-opt (%) | **66.0**± 5.7 | 51.1± 0.0 | 65.2± 8.4 |

Table 4: Experimental results for the discriminative models.

### 6.2 Conditional: Arbitrary Bregman Divergence

**Algorithms.** Our method (cvxCond) first optimized (15) over $M \in \mathcal{M}_2$ using Algorithm 1. Then similar to §6.1, the optimal $M$ was rounded by spectral clustering (10 repeats). Here subsequent re-optimization (based on local optimization) was performed on the objective $D_{F^*}(YB, f(X))$. The competing algorithm, altCond, optimizes this objective by alternating with 30 random initializations of $Y$.

**Results.** The results in Table 3 are organized in the same manner as Table 2. Here it can be observed that for all data sets and transfer functions, cvxCond with SC rounding and reoptimization yields lower optimal objective value and higher accuracy than altCond (except Diabetes/sigm). Moreover, the objective values also exhibits lower standard deviation than altCond, which suggests that the value regularization scheme helps stabilize the reoptimization. Finally note the accuracy of cvxCond with rounding is already comparable with that of altCond on most data sets.

### 6.3 Discriminative Models

**Algorithms.** Our method (cvxDisc) optimized (28) over $M \in \mathcal{M}_2$ by solving (29). We also tested on the algorithms of (Joulin & Bach, 2012) and (Guo & Schuurmans, 2007), which we refer to as JB and GS respectively. The result of all the three methods were rounded by spectral clustering, then used to initialize a local re-optimization over $D_F(X, Y\Upsilon)$. Since the discriminative model is logistic, we used the sigmoid transfer in $D_F$ only.

**Results.** According to Table 4, it is clear that even without reoptimization, cvxDisc after rounding already achieves higher or comparable accuracy to both JB and GS in all cases. Further improvements are obtained by reoptimization. Regarding the run time for solving the respective convex relaxations, cvxDisc is at least 10 times faster than both JB and GS. This confirms the computational advantage of our primal reformulation (28), compared to other implementations of convex relaxation.

### 6.4 Joint Generative Models

**Algorithms.** Our proposed method, cvxJoint, optimizes (32) over $M \in \mathcal{M}_2$ by solving (33). As before, we rounded the optimal $M$ by spectral clustering, and used the $Y$ to initialize local reoptimization of the joint likelihood $-\mathbf{1}'Y\mathbf{w} + tg(\mathbf{w}) + L(YB)$.

We compared the results to those of three soft generative models. The standard soft EM (Banerjee et al., 2005, Algorithm 3) was randomly reinitialized 20 times. The other two algorithms are LG (Lashkari & Golland, 2007), and NB[6] (Nowozin & Bakir, 2008). Since they do not directly control the number of clusters, we tuned their parameters so that the resulting number of cluster is $d$, or a little higher than $d$ which could be truncated based on the cluster prior.

**Results.** Since joint models also learn a cluster prior, accuracy can take two forms. The hard accuracy is computed by $\mathrm{argmax}_y\, p(y|\mathbf{x}_i) = \mathrm{argmax}_y\, p(y)p(\mathbf{x}_i|y)$ in the case of soft EM, LG, and NB. Our model outputs a hard accuracy by locally reoptimizing the joint likelihood. For all methods, we define the soft accuracy based on the posterior distribution: $\max_\pi \mathbb{E}_{Y \sim p(Y|X)}[\mathrm{Accuracy}(Y, \pi(Y^*))]$, where $Y^*$ is the ground truth label and $\pi$ is a matching between the cluster and label.

As can be observed from Table 5, cvxJoint with rounding and reoptimization achieves superior or comparable performance to the competing algorithms in most cases (except three settings in Balanced and one each in Yale and Diabetes), both in terms of hard *and* soft accuracy.

## 7 Conclusion

In this paper we constructed convex relaxations for clustering with Bregman divergences. Using normalized equivalence relations, we also designed efficient algorithms for optimizing the models. For future work, it will be interesting to extend these approaches to generative soft clustering, and further scale up the optimization to large applications.

---

[6] http://www.nowozin.net/sebastian/infex. Since their approach relies heavily on the Gaussian model, we put NA in the corresponding cells in Table 5.

|  | linear | | sigmoid | |
|---|---|---|---|---|
|  | acc(%) | soft acc(%) | acc(%) | soft acc(%) |
| | Spam E-mail | | | |
| cvxJoint1 | 55.7±1.9 | 55.9±1.4 | 62.6±9.0 | 67.7±11.0 |
| cvxJoint2 | 60.5±0.0 | 60.5±0.0 | **81.5**±16.4 | **79.2**±15.1 |
| softEM | 60.5±0.0 | 54.5±2.6 | 58.2±7.4 | 52.9±2.0 |
| LG | 60.0 | 0.1 | 40.6 | 1.8 |
| NB | **60.5** | 51.4 | NA | NA |
| | ORL | | | |
| cvxJoint1 | **61.0**±1.3 | 52.6±1.5 | **63.0**±2.3 | 58.6±1.8 |
| cvxJoint2 | 55.9±1.4 | 52.8±1.2 | 58.7±2.7 | **58.7**±2.7 |
| softEM | 39.6±2.1 | 37.0±2.0 | 44.9±3.1 | 44.7±3.1 |
| LG | 40.0 | 1.9 | 36.0 | 0.5 |
| NB | 12.0 | 5.3 | NA | NA |
| | Yale | | | |
| cvxJoint1 | **47.9**±3.8 | **45.9**±3.1 | 61.9±8.3 | 55.9±1.4 |
| cvxJoint2 | 45.8±3.4 | 45.1±3.1 | 60.5±0.0 | 60.5±0.0 |
| softEM | 39.6±2.1 | 37.0±2.0 | 60.5±0.0 | 60.5±0.0 |
| LG | 35.2 | 4.8 | **66.9** | 0.1 |
| NB | 20.6 | 10.4 | NA | NA |
| | Balance | | | |
| cvxJoint1 | 50.5±2.3 | 36.3±0.7 | 51.6±2.7 | 39.5±1.2 |
| cvxJoint2 | 46.1±0.0 | 46.1±0.0 | 46.1±0.0 | **46.1**±0.0 |
| softEM | 46.1±0.0 | 38.1±2.8 | 46.1±0.0 | 39.6±0.0 |
| LG | **57.4** | 0.2 | **59.0** | 0.2 |
| NB | 54.2 | **54.7** | NA | NA |
| | Breast Cancer | | | |
| cvxJoint1 | **71.0**±11.9 | 56.9±4.7 | **70.9**±13.0 | 63.9±8.1 |
| cvxJoint2 | 65.5±0.0 | **65.5**±0.0 | 65.5±0.0 | **65.5**±0.0 |
| softEM | 65.5±0.0 | 57.7±4.5 | 65.5±0.0 | 55.5±5.4 |
| LG | 61.8 | 0.1 | 65.5 | 0.1 |
| NB | 69.8 | 50.3 | NA | NA |
| | Diabetes | | | |
| cvxJoint1 | 56.0±2.6 | 53.6±2.5 | 57.5±5.5 | 57.6±5.6 |
| cvxJoint2 | **65.1**±0.0 | **65.1**±0.0 | 62.0±3.3 | 62.6±2.6 |
| softEM | 65.1±0.00 | 57.6±4.6 | **65.1**±0.0 | 57.4±5.2 |
| LG | 56.8 | 0.1 | 58.5 | 0.1 |
| NB | **65.1** | 60.2 | NA | NA |
| | Heart | | | |
| cvxJoint1 | **63.0**±6.4 | 53.3±1.8 | 63.0±7.4 | 61.0±6.2 |
| cvxJoint2 | 55.6±0.0 | **55.5**±0.0 | **64.0**±7.5 | **61.3**±7.1 |
| softEM | 55.6±0.0 | 51.7±1.6 | 55.6±0.0 | 52.7±0.0 |
| LG | 57.4 | 0.4 | 55.2 | 0.4 |
| NB | 55.6 | 53.0 | NA | NA |

Table 5: Experimental results for the joint generative model. Here cvxJoint1 is cvxJoint followed by SC rounding, whereas cvxJoint2 uses additional re-optimization. Best results in **bold**.

**Acknowledgements**

This research is supported by AICML and NSERC. We thank Junfeng Wen and Yaoliang Yu for their helpful discussions and early assistance. NICTA is funded by the Australian Government as represented by the Department of Broadband, Communications and the Digital Economy and the Australian Research Council through the ICT Centre of Excellence program.